\pdfoutput=1

\documentclass[11pt]{article}

\usepackage{acl}
\usepackage{amsmath}
\usepackage[ruled]{algorithm2e}
\usepackage{times}
\usepackage{latexsym}
\usepackage{svg}
\usepackage[T1]{fontenc}

\usepackage[utf8]{inputenc}

\usepackage{microtype}

\usepackage{inconsolata}
\usepackage{ragged2e}
\usepackage{mdframed}
\usepackage{graphicx}
\usepackage{subcaption}
\usepackage{longtable}
\usepackage{multirow}
\usepackage[table]{colortbl}%
\usepackage{caption}
\usepackage{booktabs}

\newcolumntype{P}[1]{>{\centering\arraybackslash}p{#1}}

\newcommand\prompt[1]{$\mathcal{D}_\textrm{{#1}}$}

\newcommand{\best}[1]{\textbf{#1}}

\newcommand{\factscore}{\textsc{FActScore}}
\newcommand{\factscoresentencecontext}{\textsc{DecompScore}}
\newcommand{\propsegment}{\textsc{PropSegmEnt}}
\newcommand{\wice}{\textsc{WiCE}}
\newcommand{\faithscore}{\textsc{FaithScore}}
\newcommand{\ourmethod}{\prompt{R-ND}}

\DeclareUnicodeCharacter{01B0}{01AF}
\DeclareUnicodeCharacter{0300}{00D2}
\DeclareUnicodeCharacter{01A1}{01A1}

\title{A Closer Look at Claim Decomposition}

\author{Miriam Wanner$^*$, Seth Ebner$^*$, Zhengping Jiang,\\ {\bf Mark Dredze}, {\bf Benjamin Van Durme}\\[1em] Johns Hopkins University \\[1em] 
\texttt{\{mwanner5,seth,zjiang31,mdredze,vandurme\}@jhu.edu} \\ [1em] }

\begin{document}
\maketitle
{
\def\thefootnote{*}\footnotetext{Equal contribution}
}
\begin{abstract}

As generated text becomes more commonplace, it is increasingly important to evaluate how well-supported such text is by external knowledge sources. Many approaches for evaluating textual support rely on some method for decomposing text into its individual subclaims which are scored against a trusted reference. We investigate how various methods of claim decomposition---especially LLM-based methods---affect the result of an evaluation approach such as the recently proposed \factscore{}, finding that it is sensitive to the decomposition method used. This sensitivity arises because such metrics attribute overall textual support to the model that generated the text even though error can also come from the metric's decomposition step. To measure decomposition quality, we introduce an adaptation of \factscore{}, which we call \factscoresentencecontext{}. We then propose an LLM-based approach to generating decompositions inspired by Bertrand Russell's theory of logical atomism and neo-Davidsonian semantics and demonstrate its improved decomposition quality over previous methods.

\end{abstract}

\section{Introduction}

\begin{figure}[ht!]
    \centering
    \includegraphics[width=0.87\linewidth]{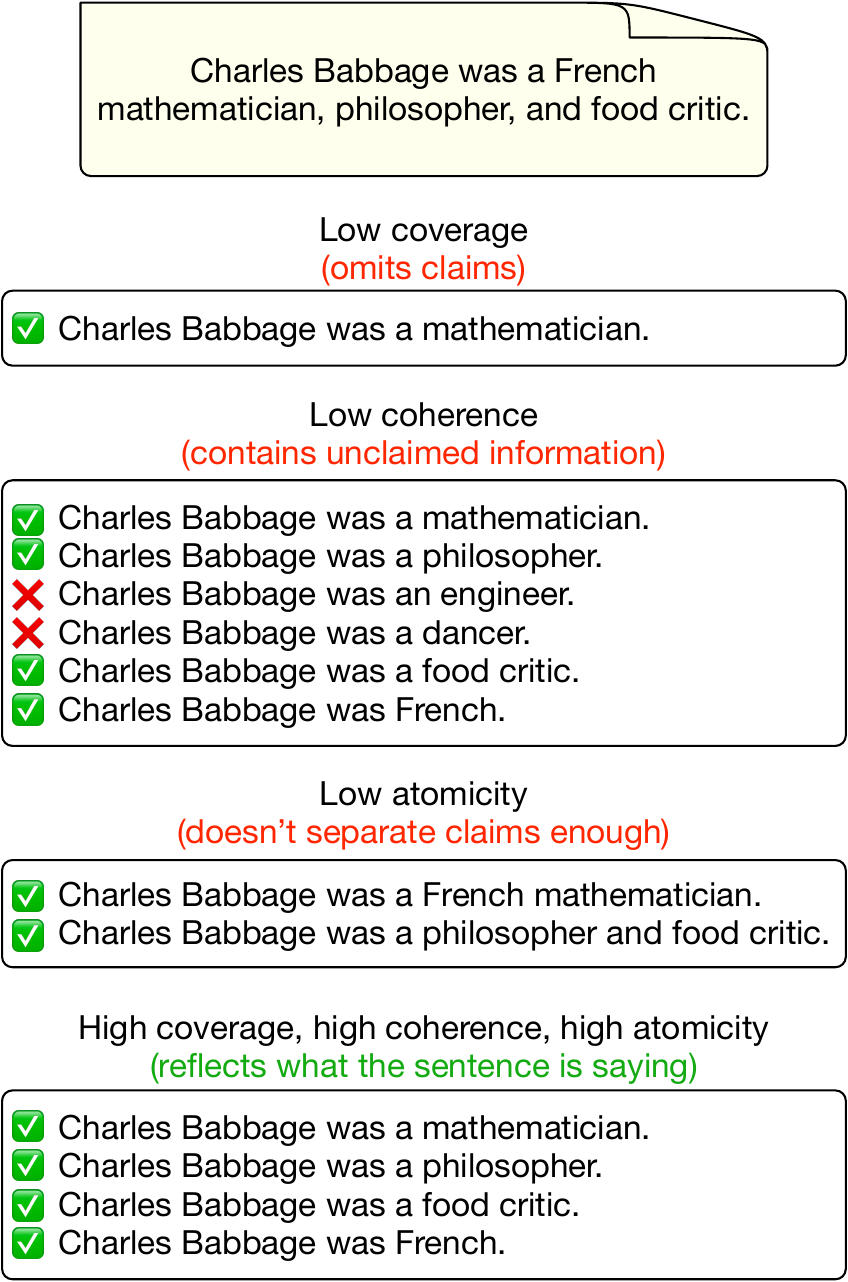}
    \caption{Modes of claim decomposition. The extent to which textual support can be determined depends on how the generated text (yellow box) is decomposed into its subclaims (white boxes). Higher quality decompositions enable more complete identification of discrepancies between generated text and a reference (not shown), which consequently increases the reliablility of the downstream textual support metric. Checks and Xs denote that the statement is claimed or is not claimed, respectively, by the generated text.}
    \label{fig:decomposition-modes}
\end{figure}

Recent work uses claim decomposition to determine how well supported a claim is for applications in factual precision of generated text \cite{min2023factscore}, entailment of human generated text \cite{kamoi-etal-2023-wice,chen-etal-2023-propsegment}, and claim verification \cite{chen2023complex,li2023self,milbauer-etal-2023-newssense}, with similar ideas going back decades \cite{hickl-bensley-2007-discourse}. In each of these cases, a claim is decomposed into natural language subclaims,\footnote{The terms ``atomic fact'' and ``atomic proposition'' are also used for similar concepts.} typically using a large language model (LLM), and each subclaim is then scored or aligned to information from external sources using a task-specific metric. %
Claim decompositions with various characteristics are shown in \autoref{fig:decomposition-modes}.

Evaluating subclaims individually, as opposed to the entire claim at once, we can assign partial credit to a claim (e.g., for partial support), identify which parts of the claim differ from reference texts (such as a retrieved or pre-specified document or passage), and more easily identify relevant source material for each part of the claim. Claims can come from human-authored text based on cited documents \cite{kamoi-etal-2023-wice,chen-etal-2023-propsegment,chen2023sub} or from machine-generated text based on dynamically provided grounding text or text observed during pre-training \cite{min2023factscore}.

Since claim decomposition determines the number and scope of each evaluated subclaim, any analysis or resulting metric will be inherently tied to the decomposition method.
Nevertheless, prior work has left decomposition itself largely untested. How do different decomposition strategies affect downstream analysis? What are their qualitative and quantitative similarities and differences?

We show that a downstream metric of textual support such as \factscore{} \cite{min2023factscore} is sensitive to the decomposition method it uses (\autoref{fig:factscore_scatterplot}). While \factscore{} aims to measure the factual precision of generated text, the number and nature of the subclaims it evaluates from that text depend entirely on the metric's claim decomposition method. The higher the quality of the decomposition method, and the better we understand its characteristics, the more we can attribute the factual precision that \factscore{} aims to measure to the text generation model rather than to artifacts of the decomposition.

Finding that the method of claim decomposition matters, we introduce \factscoresentencecontext{}, an adaptation of \factscore{} that measures decomposition quality, an important step in determining the reliability of the downstream metric. \factscoresentencecontext{} measures the number of subclaims supported by the original claim that was decomposed. Because a decomposition with high atomicity and coverage will have more subclaims than a decomposition that doesn't, we then favor the decomposition method with the greatest \factscoresentencecontext{}, especially when coupled with qualitative evidence of high atomicity and coverage.

With a way to compare decomposition methods in hand, we propose an LLM-based decomposition approach inspired by Bertrand Russell's theory of logical atomism and neo-Davidsonian semantics. Our approach gives far more subclaims than other methods while maintaining high coherence\footnote{We use the term ``coherence'' to denote whether a subclaim accurately reflects what is stated in the original claim.} with the claim being decomposed, and thus results in greater confidence in the entire pipeline for evaluating the level of textual support.

Our contributions are:
\begin{enumerate}
    \item Empirical evidence that the method of claim decomposition affects a downstream metric of textual support;
    \item Quantitative and qualitative comparisons of claim decomposition methods;
    \item A method for claim decomposition inspired by philosophical and semantic theories that outperforms previous methods.
\end{enumerate}

\begin{table*}[h]
\small
\centering
\begin{tabular}{m{0.1\textwidth}m{0.28\textwidth}m{0.06\textwidth}m{0.06\textwidth}m{0.16\textwidth}m{0.16\textwidth}}
    \toprule
    \multicolumn{1}{c}{\multirow{2}{*}{Method}} & \multicolumn{1}{c}{\multirow{2}{*}{Instruction}} & \multicolumn{4}{c}{In-Context Examples}\\
    
    & & \multicolumn{1}{c}{Static} & \multicolumn{1}{c}{Retrieved} & \multicolumn{1}{c}{Sentences} & \multicolumn{1}{c}{Decompositions}\\
    \midrule
    \multicolumn{1}{c}{\prompt{\factscore{}}} & {\scriptsize ``Please breakdown the following sentence into independent facts:'' \cite{min2023factscore}}   & \multicolumn{1}{c}{7} & \multicolumn{1}{c}{1} & \multicolumn{1}{c}{\citet{min2023factscore}} & \multicolumn{1}{c}{\citet{min2023factscore}} \\ 
    \rowcolor{gray!10} \multicolumn{1}{c}{\prompt{\wice{}}} & \scriptsize ``Segment the following sentence into individual facts:'' \cite{kamoi-etal-2023-wice} & \multicolumn{1}{c}{6} & \multicolumn{1}{c}{0} & \multicolumn{1}{c}{\citet{kamoi-etal-2023-wice}} & \multicolumn{1}{c}{\citet{kamoi-etal-2023-wice}}  \\
    \multicolumn{1}{c}{\prompt{Chen et al.}} & \scriptsize ``Given the following sentence, tell me what claims they are making. Please split the sentence as much as possible, but do not include information not in the sentence:'' \cite{chen2023sub} & \multicolumn{1}{c}{7} & \multicolumn{1}{c}{1} & \multicolumn{1}{c}{\citet{min2023factscore}} & \multicolumn{1}{c}{\citet{min2023factscore}} \\
    \rowcolor{gray!10} \multicolumn{1}{c}{\prompt{CoNLL-U}} & {\scriptsize ``The sentence below is given in CoNLL-U format. Word lines contain the annotation of a word/token/node in 10 fields separated by single tab characters. Sentences consist of one or more word lines. Please break down the following sentence given in CoNLL-U format into independent facts:''} & \multicolumn{1}{c}{1} & \multicolumn{1}{c}{1} & \citet{min2023factscore} + CoNLL-U Parse & \multicolumn{1}{c}{\citet{min2023factscore}} \\
    \multicolumn{1}{c}{\ourmethod{}} & {\scriptsize ``Please decompose the following sentence into individual facts:''}   & \multicolumn{1}{c}{7} & \multicolumn{1}{c}{1} & \multicolumn{1}{c}{\citet{min2023factscore}} & \multicolumn{1}{c}{\textbf{Manual (ours)}}\\
    \bottomrule
\end{tabular}
\caption{Summary of LLM prompted claim decomposition methods (method names are prefixed with \prompt{} for ``decomposer''). The prompt given to the LLM is a concatenation of the instruction, statically and dynamically selected in-context examples, and the sentence to be decomposed. The in-context decomposition examples used in our approach (\ourmethod{}) are based on Russellian and neo-Davidsonian theories (\S\ref{sec:russell theory}).}
\label{tab:prompt specs}
\end{table*}

\section{Localized Textual Support}
\label{sec:textual_support_measures}

\factscore{} \cite{min2023factscore} and \wice{} \cite{kamoi-etal-2023-wice} are representative examples of current LLM-based approaches for determining support for particular claims for different downstream use cases. Broadly, methods of this type decompose a claim into its subclaims, evaluate each subclaim for its level of support based on external sources, and then aggregate results to give a single score or label for the entire claim. Since each subclaim is evaluated, we get a localized view of which parts of the claim are supported. The more atomic the subclaims are, the more precisely we can localize the information in the claim that differs from a trusted reference. Since these approaches rely on decomposition, the better the decomposition method the more reliable the results.

\factscore{} \cite{min2023factscore} measures factual precision of model-generated text with respect to a knowledge source. %
A generated passage is split into sentences, which are decomposed into subclaims by an LLM. The percentage of subclaims supported by a retrieved knowledge source (e.g., Wikipedia excerpts) is the \factscore{} for the passage. %
\faithscore{} \cite{jing2023faithscore} takes a similar approach for evaluating the outputs of vision-language models, in which the knowledge source against which the subclaims are evaluated is an image. They additionally require that the subclaims fit into certain domain-specific categories such as color and count. %

The \wice{} dataset \cite{kamoi-etal-2023-wice} contains annotations for whether subclaims in human-written text are supported, partially supported, or not supported by external reference documents, from which claim-level support labels are derived. %
\citet{kamoi-etal-2023-wice} also apply their LLM-based Claim-Split approach to entailment classification, in which entailment scores for each subclaim are aggregated to give an entailment score for the whole claim. %

\section{Evaluating Decomposition Quality}
Previous work on evaluating the veracity of generated text attributes the resulting score to the quality of the generation, overlooking the role of metric's decomposition step. However, higher quality decompositions mean that we can more reliably measure the quality of the generation.
Depending on the characteristics of the decomposition method (e.g., how atomic its decompositions are), a metric like \factscore{} can change for the same underlying generated text (\autoref{fig:factscore_scatterplot}). Furthermore, \factscore{} implicitly assumes complete and coherent decompositions. However, the decomposition step can introduce unclaimed information or omit existing (possibly incorrect) claims, which introduces measurement error into \factscore{}.

\subsection{Qualitative Evaluation}
What makes a decomposition higher quality? The subclaims must be faithful to the original claim. In other words, they must cohere with (are supported or entailed by) the original claim.\footnote{In contrast to the coherence theory of truth, the correspondence theory deems a statement to be true if it matches a situation in reality. It is not in the purview of a decomposition model to determine whether a claim agrees with a knowledge source; that is the purpose of the validator. In other words, the validator is the ``fact checker''. A validator that appeals to a knowledge source is actually following a coherence theory of truth (where the given set of statements is the information contained in the knowledge source). The validator's adherence to a coherence theory of truth is apparent if we consider a case in which the subclaims are not grounded in reality but rather derived from a work of fiction. We can judge a statement like ``Sherlock Holmes lives at 221B Baker Street'' to be true even though it is false in reality.} To be of the greatest use for localizing discrepancies with a trusted reference, the subclaims should cover all parts of the claim and also be as atomic as possible. Different methods decompose claims to various degrees, with some methods producing more or fewer subclaims. We explore these various characteristics across decomposition methods in \S\ref{sec:qualitative_analysis}.

\subsection{Quantitative Evaluation: \factscoresentencecontext{}} \label{sec:d-factscore}

We develop a measure of decomposition method quality by utilizing the same procedure as \factscore{}, namely using an LLM to assign a binary judgment of support for every subclaim. Rather than providing an external knowledge source as context for the validator, we provide the original sentence that was decomposed, thus identifying the subclaims that are supported by the original sentence.

The \factscoresentencecontext{} of a decomposition method is the average number of supported subclaims per passage produced by that decomposition method.
This metric indicates which method generates the most subclaims that cohere with the sentence being decomposed. For example, if a text is decomposed into a large number of subclaims but \factscoresentencecontext{} is low, we can infer that the subclaims produced by the decomposition method are not of good quality.
The optimal value of \factscoresentencecontext{} for a particular passage is difficult to determine because we do not have a set of reference decompositions, but in general, methods that produce decompositions with high atomicity and coverage will achieve higher \factscoresentencecontext{}.

Entailment is another notion of coherence that could be used to evaluate whether a subclaim is a valid part of the decomposition. In practice, we find high correlation (\autoref{fig:nli-entailment} in \autoref{app:nli_entailment}) between \factscoresentencecontext{} and the average number of subclaims entailed by the original claim using a strong natural language inference (NLI) model \cite{nie-etal-2020-adversarial}.\footnote{\url{https://huggingface.co/ynie/roberta-large-snli_mnli_fever_anli_R1_R2_R3-nli}}

\section{Methods of Claim Decomposition} \label{sec:methods}

We study three types of claim decomposition methods, which are discussed below.

\subsection{LLM prompting}\label{subsec:llm_prompting}

Much of the recent work for claim decomposition utilizes a prompted LLM-based method, typically with in-context example decompositions \cite{min2023factscore, kamoi-etal-2023-wice, chen2023sub, jing2023faithscore,mohri2024language}. The in-context examples can be dynamically selected using a retrieval model \cite{min2023factscore}. We use three instructions from prior work \cite{min2023factscore, chen2023sub, kamoi-etal-2023-wice} and one of our own, with various static and retrieved in-context examples. Notably, our approach uses manually decomposed in-context examples based on philosophical and linguistic theories, which are discussed in \S\ref{sec:russell theory}. The approaches' configurations are outlined in \autoref{tab:prompt specs}.

The LLM prompting approach is flexible and unstructured, allowing for the generation of arbitrary text. This text generation nature of LLMs produces fluent natural language decompositions by incorporating words outside the original sentence (in contrast to, e.g., \propsegment{} \cite{chen-etal-2023-propsegment}), but this also permits hallucinations and forces us to relinquish control over the model's outputs due to the large output space. We can adapt the instructions and in-context examples to encourage certain characteristics in the output (such as coherence and atomicity), but ultimately there is no mechanism to guarantee they are reflected in the output. However, in-context examples that are dynamically chosen based on high similarity with the claim to be decomposed could encourage similar styles of decomposition, which may provide some amount of controllability. %
A simple prompt-in, subclaims-out interface also avoids issues of parsing into and generating out of an explicit intermediate semantic representation, designing such a representation in the first place, and overcoming any structural weaknesses in such a representation. %

\subsection{Shallow semantic parsing}
Rather than rely on an LLM, we can use a semantic analysis of the text to decompose claims. We use PredPatt \citep{white-etal-2016-EMNLP,zhang-etal-2017-IWCS}, a system for extracting predicate-argument sub-structures from a syntactic dependency parse. We take these sub-structures as representing the propositional content of subclaims. \citet{goyal-durrett-2020-evaluating} use similar intuitions about a correspondence between syntactic dependency arcs and semantic units to decompose a claim based on the arcs in a dependency parse.

The resulting subclaims contain only words from the original sentence, and are often not grammatical sentences.\footnote{PredPatt can add short strings like ``is/are" and ``poss" to indicate being and possession, respectively, but these additions do not make the propositions fluent.} The subclaims in a valid decomposition should be full sentences in order to be validated by \factscoresentencecontext{} and \factscore{}, and for this reason, we use an LLM (\texttt{gpt-3.5-turbo-instruct}) to convert the PredPatt parses into fluent, natural language. Details are given in \autoref{sec:model specs}. Although the resulting strings are often full grammatical sentences, the LLM does not guarantee this behavior.\footnote{A model for determining grammatical acceptability could be included in this approach to filter out ungrammatical strings or send them back for rewriting \cite{warstadt-etal-2019-neural}.}

\subsection{LLM prompting with parse}\label{subsec:prompt_w_parse}

Combining syntactic structure with the flexibility of text generation could support a more grounded decomposition from an LLM. We use an LLM prompting method, but this time supplied with a parsed version of the original sentence. We use Trankit \citep{nguyen-etal-2021-trankit}, a state-of-the-art dependency parser, to obtain dependency parses \cite{ud25} of each claim as well as each in-context learning example. Because CoNLL-U formatted parses \cite{nivre-etal-2017-universal} are token-heavy, fewer in-context examples are provided. Prompt details can be found in \autoref{tab:prompt specs}.

This method inherits the fluency and flexibility of LLM prompting while grounding the LLM's response in a syntactic analysis, resulting in (hopefully) a higher quality decomposition. While we hope the added structure imposes controllability, LLMs can still generate subclaims that do not cohere with the original claim. %

\section{Russellian and Neo-Davidsonian decomposition}\label{sec:russell theory}
The notion of claim decomposition has roots in the philosophical literature. We draw inspiration from Bertrand Russell's theory of logical atomism for how claims should be decomposed into their atomic components.

Russell defines atomic facts as properties of individuals or relations between individuals from which all other facts are composed \cite{russell1918philosophylecture2}.\footnote{Ludwig Wittgenstein theorizes a similar idea of elementary propositions that assert atomic ``states of affairs''. On the whole, we find Wittgenstein's theory to be less actionable than Russell's. Incidentally, Wittgenstein later abandoned this theory in part due to the color exclusion problem, which we avoid by not requiring independence of subclaims, instead requiring only that each subclaim is claimed by the sentence.}\textsuperscript{,}
\footnote{A note on terminology: For Russell, ``facts'' are ``the kind of thing that makes a proposition true or false'' \cite{russell1918philosophylecture1}, and for Wittgenstein they are states of affairs. In both cases, they are not propositions but rather conditions of the world. Russell and Wittgenstein use the terms ``atomic proposition'' and ``elementary proposition'', respectively, to refer to the corresponding truth function or expression of an atomic fact. The NLP literature uses the term ``atomic fact'' to mean the corresponding proposition, typically written in natural language.} We take individuals to be entities and eventualities mentioned in the sentence. This kind of Russellian analysis accords with neo-Davidsonian analysis \cite{castaneda1967,Parsons1990-PAREIT} (building on \citet{Davidson1967-DAVTLF}), in which the logical form of a sentence is decomposed fully to a conjunction of unary predicates (akin to properties of individuals) and binary predicates (akin to relations between individuals).

We manually decompose the 21 in-context examples from \citet{min2023factscore} into lists of such Russellian atomic propositions that we further decompose following neo-Davidsonian intuitions into unary and binary relations to obtain the smallest units that are claimed in each sentence: each subclaim designates a property of an individual or a relation between two individuals.\footnote{We do not include existence as a property of entities. Consider the  sentences: ``Allan Pinkerton was a detective who worked in the United States.'' and ``Sherlock Holmes was a detective who worked in London.'' From just the sentences alone and without external knowledge, there is no way to tell that one of these people existed and one didn't.} Our decompositions are listed in \autoref{tab:russellian_examples}. These in-context examples are retrieved in the same way as the examples are retrieved for the \factscore{} prompt.%

\section{Data}\label{sec:data}
We use the released data from \citet{min2023factscore}, which consists of biographies of 500 individuals generated from each of 12 LMs (following their notation, we call the text generation models $\textrm{LM}_{\textrm{SUBJ}}$).\footnote{GPT-4 \citep{openai2023gpt4}, ChatGPT, InstructGPT, Alpaca 7B, 13B, and 65B \citep{alpaca-2023}, Vicuna 7B and 13B \citep{vicuna2023}, \href{https://huggingface.co/databricks/dolly-v2-12b}{Dolly 12B} \citep{biderman2023pythia}, \href{https://huggingface.co/stabilityai/stablelm-tuned-alpha-7b}{StableLM-tuned-alpha 7B} \citep{alpaca-2023, vicuna2023, gpt4all}, \href{https://huggingface.co/OpenAssistant/oasst-sft-1-pythia-12b}{Oasst-pythia 12B}, and \href{https://huggingface.co/mosaicml/mpt-7b-chat}{MPT Chat 7B}.} %
We do not modify the biographies generated by \citet{min2023factscore}, nor do we generate additional ones. We treat them as static documents to investigate various decomposition methods applied to the sentences in the biographies.

\section{Experiments}

We use the data described in \S\ref{sec:data} for sentence-level decomposition with the methods outlined in \S\ref{sec:methods}. Model specifications are listed in \autoref{sec:model specs}. We evaluate using \factscoresentencecontext{} with Inst-LLAMA from \citet{min2023factscore} (which is \textsc{LLAMA} trained on Super Natural Instructions \citep{wang-etal-2022-super, touvron2023llama}) and \factscore{} with the Inst-LLAMA + retrieval + NPM setting.
In total, generating decompositions took 120 GPU-hours, computing \factscoresentencecontext{} took 250 GPU-hours, and computing \factscore{} took 450 GPU-hours, all using a Quadro RTX 6000.

\section{Results}

\begin{figure}[h]
\centering
    \includegraphics[width=0.4\textwidth]{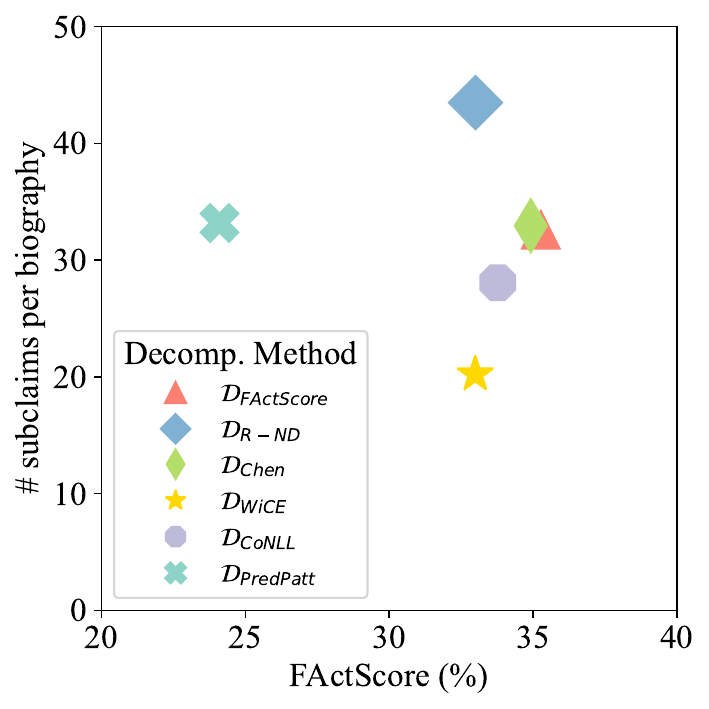}
        \caption{\factscore{} (macro-averaged across $\textrm{LM}_{\textrm{SUBJ}}$) using different decomposition methods. The same underlying set of documents is assigned different \factscore{} values depending on the decomposition method used.}%
        \label{fig:factscore_scatterplot}
\end{figure}

\begin{figure}[t]
\centering
    \includegraphics[width=0.4\textwidth]{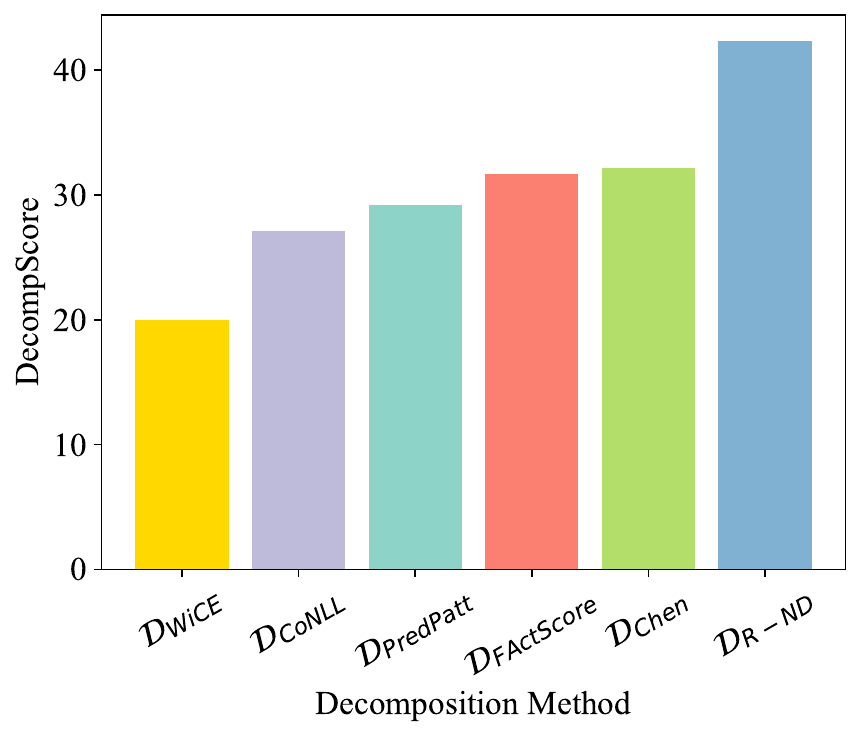}
        \caption{\factscoresentencecontext{} (macro-averaged across $\textrm{LM}_{\textrm{SUBJ}}$) of different decomposition methods. A higher \factscoresentencecontext{} is better.} %
    \label{fig:d-factscore_scatterplot}
\end{figure}

\factscoresentencecontext{} results are shown in \autoref{fig:d-factscore_scatterplot}, with full results in \autoref{tab:filtered subclaims} (\autoref{sec:full_results}). \ourmethod{} attains the highest \factscoresentencecontext{} (i.e., highest average number of supported subclaims per biography) with 42.3, followed by \prompt{Chen et al.} and \prompt{\factscore{}}, both with around 32. \prompt{\wice{}} produces the fewest average supported subclaims, with a \factscoresentencecontext{} of 20.0, less than half that of \ourmethod{}. The \factscoresentencecontext{}s of \prompt{PredPatt} and \prompt{CoNLL-U} fall between \prompt{\wice{}} and \prompt{\factscore{}}, with \prompt{PredPatt} achieving a slightly higher \factscoresentencecontext{} (29.2) than \prompt{CoNLL-U} (27.1).

\factscore{} results are shown in \autoref{fig:factscore_scatterplot}, with full results in \autoref{tab:factscore} and \autoref{fig:scatter_all_factscore} (\autoref{sec:full_results}). Undesirably, the \factscore{} values vary based on the decomposition method used. %

\subsection{Qualitative Analysis}\label{sec:qualitative_analysis}

We analyze all decomposition methods on two sentences generated by GPT-4: one about Alfred Hitchcock and one about John Nash.\footnote{``Alfred Hitchcock passed away on April 29, 1980, in Bel-Air, California, leaving behind a rich legacy of suspenseful and thrilling films that continue to captivate and inspire audiences and filmmakers alike.'' and ``Nash demonstrated a natural aptitude for mathematics from a young age and earned his bachelor's and master's degrees in mathematics from the Carnegie Institute of Technology (now Carnegie Mellon University) in 1948.''} The decompositions, alongside our own manual decompositions, are shown in \autoref{tab:hitchcock_decompositions} and \autoref{tab:nash_decompositions} in \autoref{appendix:decomposition_outputs}. The evaluation criteria we use are coherence to the original sentence, coverage of the information claimed, and atomicity.

We observe that for the sentence about Alfred Hitchcock (\autoref{tab:hitchcock_decompositions}), no decomposition method separates the date into month, day, and year or the location into city and state. No method generates the subclaim ``Alfred Hitchcock passed away'', opting to always include the date or location. Additionally, no method outputs all four combinations arising from the conjunction of ``captivate'' and ``inspire'' with ``audiences'' and ``filmmakers''. \ourmethod{} is the only method to separate ``suspenseful'' from ``thrilling''; every other method keeps them as one unit. Similarly, many methods keep ``captivate and inspire'' as one unit; \ourmethod{} and \prompt{\factscore{}} are the only ones to always split this conjunction. 

We see that for the sentence about John Nash (\autoref{tab:nash_decompositions}), \ourmethod{}, \prompt{\factscore{}}, and \prompt{Chen et al.} all output a large number of subclaims. However, many of the subclaims generated by \prompt{\factscore{}} and \prompt{Chen et al.} incrementally add information to their other subclaims, which makes them non-atomic. This incrementality makes it more difficult to localize errors in the original claim because the textual support of the new information in the subclaim undesirably depends on the re-used information also being supported.\footnote{This behavior of incrementally adding information can be expected given that it occurs in the in-context examples used by \citet{min2023factscore}.} All methods except for \prompt{\wice{}} generate non-atomic subclaims that combine Nash's bachelor's and master's degrees. \ourmethod{}, \prompt{CoNLL-U}, and \prompt{PredPatt} mention the degrees without the additional information that they were for mathematics, which increases atomicity; the other methods describe them always as ``degree[s] in mathematics''.

In our experiments, \prompt{\factscore{}} and \prompt{Chen et al.} use the same in-context examples with slightly different instructions and generate similar decompositions on the two sentences (identical decompositions on the Nash sentence). This behavior suggests that the in-context examples influence the decomposition more than the instruction does.

\paragraph{Takeaways} For both sentences, we observe that many subclaims in our manual decompositions are missed by the decomposition methods, but the methods with the most coverage are \ourmethod{}, \prompt{Chen et al.}, \prompt{\factscore{}}, and \prompt{\wice{}}. All methods but \prompt{PredPatt} have perfect coherence for both sentences. In general, we observe that \prompt{\wice{}} has low atomicity,\footnote{The instructions given to annotators for evaluating \wice{}'s Claim-Split decomposition method include an example that explicitly states that one of its subclaims can be further decomposed but to ignore that issue, which suggests atomicity is not prioritized in that method.} as does \prompt{CoNLL-U} because it does not split conjunctions. \prompt{PredPatt} exhibits many issues: its subclaims are not atomic, often not fluent (despite using an LLM to make them more fluent), and not coherent with the original claim (e.g., ``The bachelor possessed a master's degree'').

\subsection{Quantitative Analysis}

Even though all decomposition methods are run on the same set of static biographies, they differ in \factscore{} and number of subclaims generated (averaged over $\textrm{LM}_{\textrm{SUBJ}}$: \autoref{fig:factscore_scatterplot}, per $\textrm{LM}_{\textrm{SUBJ}}$: \autoref{tab:factscore}). This finding indicates that \factscore{} is sensitive to the method of decomposition that is used. The most reliable estimate of the generated text's ``true'' factual precision is the \factscore{} achieved by the highest quality decomposition method.

We hypothesize that \prompt{PredPatt}'s \factscore{} is low because it produces subclaims not likely to be supported by the external knowledge source,\footnote{For example, the mention of ``civil rights'' results in the subclaim ``Rights are civil'', which is likely not explicitly asserted in the retrieved Wikipedia passages.} while also being constrained to using only the words in the sentence and missing implicit subclaims not extractable as predicate-argument structures from the dependency parse. Additionally, only 86\% of the subclaims it produces are supported by the original claim (\autoref{tab:d-factscore} in \autoref{sec:full_results}), which agrees with our previous observation that its outputs have low coherence.

\prompt{\factscore{}} and \prompt{Chen et al.} both achieve a \factscoresentencecontext{} around 32, and since they use the same in-context examples in our experiments, this further suggests that the decompositions are robust to the wording of the instruction in the prompt. Additionally, the similarity of the configuration of \ourmethod{} to those of \prompt{\factscore{}} and \prompt{Chen et al.} suggests that it is the manually decomposed in-context examples used in \ourmethod{} that are responsible for its higher \factscoresentencecontext{}.

Because the in-context examples seem to have a larger effect on the decompositions than the instructions do and because we provide fewer examples in \prompt{CoNLL-U} due to the large token count of the parses, we evaluate the effect on decomposition of the number of in-context examples given. We use the same prompt specifications as in \prompt{\factscore{}} in \autoref{tab:prompt specs}, but use the same number of static examples as in \prompt{CoNLL-U} (one). We find that using fewer examples produces around the same number of subclaims (+1.3 subclaims on average), and achieves similar \factscoresentencecontext{} (-0.69\%) and \factscore{} (+0.06\%). Overall, using fewer in-context examples does not appear to have much impact on either decomposition quality or factual precision.

When evaluating \factscore{} on only the \emph{supported} subclaims (as determined in the calcluation of \factscoresentencecontext{}), in most cases, this subset of subclaims yields a higher \factscore{} (\autoref{tab:factscore}, \autoref{tab:filtered factscore}, \autoref{fig:scatter_all_factscore}, \autoref{fig:scatter_all_filtered_factscore} in \autoref{sec:full_results}),\footnote{There are 4 exceptions out of 84 cases, and the maximum decrease in \factscore{} is 0.2\%.} indicating that subclaims which do not cohere with the original sentence are likely also not supported by the knowledge source. Although simple, this filtering step removes potential errors introduced during decomposition. The fewest amount of subclaims (0.2 on average) are removed from \prompt{\wice{}}'s decompositions (compare \autoref{tab:filtered subclaims} and \autoref{tab:subclaims} in \autoref{sec:full_results}), indicating very high coherence, and the most are removed from \prompt{PredPatt}'s decompositions (4 subclaims per biography on average), suggesting low coherence to the original sentence. On average, 1.2 out of 43.5 subclaims are removed from \ourmethod{}'s decompositions.

To ensure that decompositions have high coherence, we recommend that subclaims produced by a decomposition method that are not supported by the original claim be filtered out (giving full coherence by construction). In doing so, unclaimed information that is introduced during the decomposition step is removed and not incorrectly attributed back to the generated text being evaluated.

\paragraph{Takeaways} 
Despite \prompt{\wice{}} having high coherence and coverage, it has the lowest \factscoresentencecontext{} because it has low atomicity, which makes it undesirable as a decomposition method for use in a localized textual support metric.

Achieving a higher \factscore{} with a particular decomposition method does not necessarily mean the decompositions are also of high quality. Although \ourmethod{} achieves lower \factscore{} than most of the other methods, it has a far higher \factscoresentencecontext{} than the other methods, which we hypothesize is due to our manually decomposed in-context examples. Such a method that produces a large number of supported subclaims that (qualitatively) have high coverage and atomicity is far more favorable in the textual support evaluation setting because it increases confidence in the results obtained from the downstream metric.

\section{Related Work}

\paragraph{Evaluation}
We evaluated decomposition methods that produce subclaims in sentential natural language, primarily by using contemporary technologies like large language models (\S\ref{sec:methods}). We review other methods of decomposition used in evaluation of textual support here.

Question answering \cite{wang-etal-2020-asking,durmus-etal-2020-feqa,scialom-etal-2021-questeval,fabbri-etal-2022-qafacteval} has been used for evaluating abstractive summarization. These methods generally ask questions only about noun phrases, require generating questions (the decomposition step), and require extracting answer spans, after which (typically lexical) heuristics determine if the answers between the summary and reference agree.%

\citet{goodrich2019assessing} evaluate summarization by extracting relation tuples from a model-generated summary which are compared to relations extracted from a ground-truth summary. \citet{fan2023evaluating} improve upon this approach by extracting fact tuples using semantic role labeling. \citet{goyal-durrett-2020-evaluating} evaluate the factuality of model-generated text by obtaining entailment labels on each arc in a dependency parse, which assumes a correspondence between syntactic dependency arcs and semantic units (the same core assumption made by PredPatt).%

In addition to evaluating \emph{whether} text is supported, there has also been work on evaluating types of textual errors \cite{pagnoni-etal-2021-understanding,devaraj-etal-2022-evaluating,mishra2024fine} and evaluating ambiguously supported claims \cite{glockner2024ambifc}. Although designed to be used at the sentence-level, such methodologies can also be applied to subclaims. For further discussion about identifying and mitigating errors in model-generated text, such as hallucinations, we refer the reader to \citet{10.1145/3571730} and \citet{ye2023cognitive}.%

\paragraph{NLI}
Decomposition is also used for sub-sentence level NLI. \propsegment{} \cite{chen-etal-2023-propsegment} identifies subclaims by marking tokens in a claim that are part of the subclaim. They use propositional-level NLI to detect hallucinations by comparing tokens in entailed and non-entailed propositions. Sub-sentence entailment judgments can also be combined to make sentence-level or paragraph-level entailment judgments more interpretable and robust \cite{stacey-etal-2022-logical,stacey2023logical,kamoi-etal-2023-wice}.

\paragraph{Fact Verification} Verifying the accuracy of statements depends on high quality decompositions to facilitate evidence retrieval. 
\citet{chen2023complex} build a system for complex claim verification by generating lists of yes/no questions that align to specific aspects of a claim. \cite{chen-etal-2022-generating} build a similar system that also asks implied subquestions. 
\citet{li2023self} and \citet{milbauer-etal-2023-newssense} align generated claims with statements in documents that entail or contradict the claim. 
Similarly, \citet{ernst-etal-2021-summary} align propositions between reference summaries and source documents---which is similar to the fact verification task. A model trained on their dataset was later used to cluster propositions in a system for multi-document summarization \cite{ernst-etal-2022-proposition}. \citet{chen2023sub} use decomposition to find matching subclaims (``atomic propositions'') across sentences to train proposition-level representations using contrastive learning. The proposition representations are used for retrieving propositions from a corpus that support a given proposition. %

\section{Conclusion}

We observe that a downstream metric of textual support, namely factual precision as measured by \factscore{}, is sensitive to the method it uses to decompose a claim into its subclaims. This finding leads us to measure decomposition quality using our proposed metric \factscoresentencecontext{} so that we can use the most appropriate decomposition method among those we consider.

We show that an LLM prompted with in-context learning examples that we manually decompose by following intuitions from logical atomism and neo-Davidsonian semantics outperforms other methods. Decompositions generated by our method contain the greatest number of subclaims supported by the original claim among the methods we consider. %
Qualitative analysis and comparison to manual decompositions demonstrate that all the decomposition methods we consider still miss subclaims and many generate non-atomic subclaims, indicating there still remains room for improvement.

\section*{Limitations}

Metrics like \factscore{} and \factscoresentencecontext{} are able to evaluate only information that is claimed in a generated text. Information relevant to an upstream query may be absent in the text, whether accidentally or intentionally, and these evaluation approaches cannot account for that.

This study is limited to the domain of entity biographies, so it is not representative of all use cases. Additionally, the data is monolingual (English), and we do not know if these results hold across other languages.

Running LLMs can be expensive. Because of this, we chose to use \textsc{llama} instead of ChatGPT as the validator, but even running that model is not financially feasible for everyone to use.

\section*{Ethics Statement}

LLMs are well-known to hallucinate information, and mitigation of hallucination is still an active area of research. Using LLMs to decompose a claim into subclaims can introduce new factual errors. Despite attempts to remove such errors (for example, by filtering out subclaims that are not supported by the original claim according to \factscoresentencecontext{}), errors can still persist. Caution must be taken when relying on text generated from a model.

\section*{Acknowledgements}
We thank Nathaniel Weir, Kate Sanders, and Anqi Liu for helpful discussions. This work has been supported in part by the U.S. National Science Foundation under grant No. 2204926.

\bibliography{anthology,custom}

\clearpage
\newpage

\appendix

\section{Full Results}\label{sec:full_results}

\begin{figure*}
    \centering
    \includegraphics[width=0.8\textwidth]{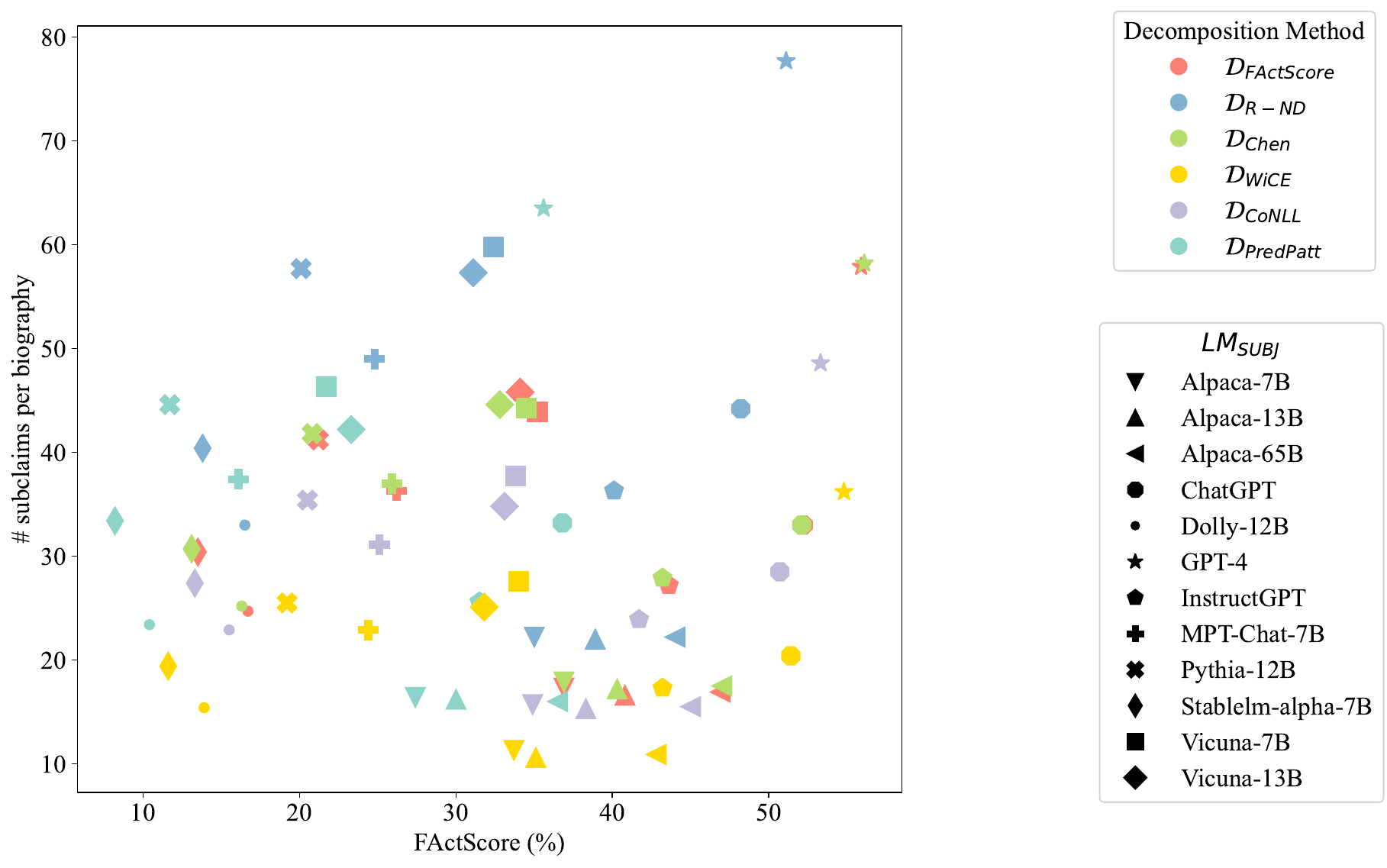}
    \caption{\factscore{} results for all claim decomposition methods and $\textrm{LM}_{\textrm{SUBJ}}$.}
    \label{fig:scatter_all_factscore}
\end{figure*}

\begin{figure*}
    \centering
    \includegraphics[width=0.8\textwidth]{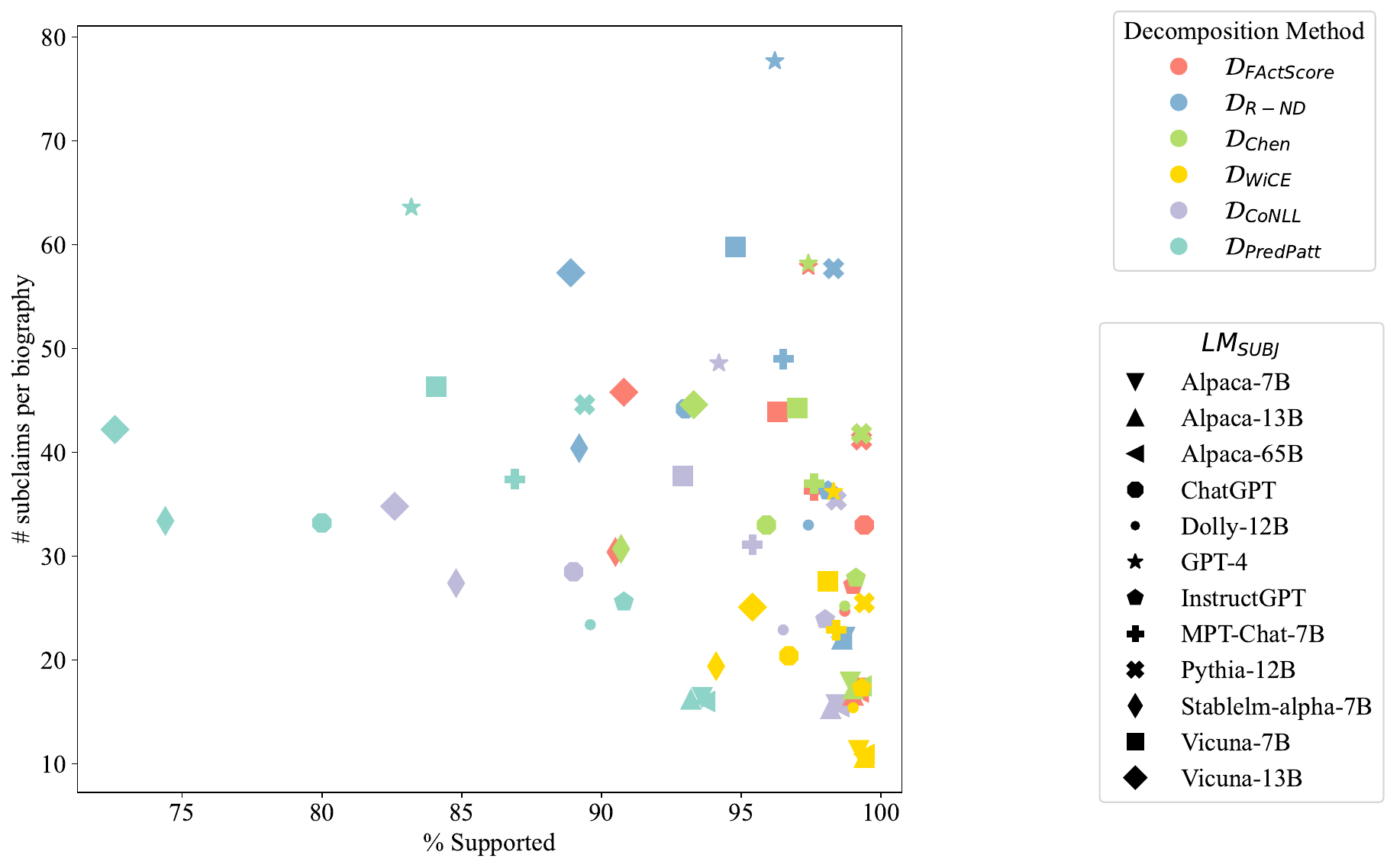}
    \caption{Percentage of subclaims that are supported by (cohere with) the original claim.} %
    \label{fig:scatter_all_d-factscore}
\end{figure*}

\factscore{} evaluation is outlined in \S\ref{sec:textual_support_measures}, and full results are reported in \autoref{tab:factscore} and \autoref{fig:scatter_all_factscore}. \factscoresentencecontext{} evaluation is discussed in \S\ref{sec:d-factscore}, and full results are reported in \autoref{tab:filtered subclaims}. Unlike \factscore{}, we do not impose a length penalty in \factscoresentencecontext{} because shorter passages naturally contain fewer subclaims. Percentages of subclaims that are judged to be supported (i.e., the coherence of each method) are shown in \autoref{tab:d-factscore} and \autoref{fig:scatter_all_d-factscore}. 

\begin{figure*}
    \centering
    \includegraphics[width=0.8\textwidth]{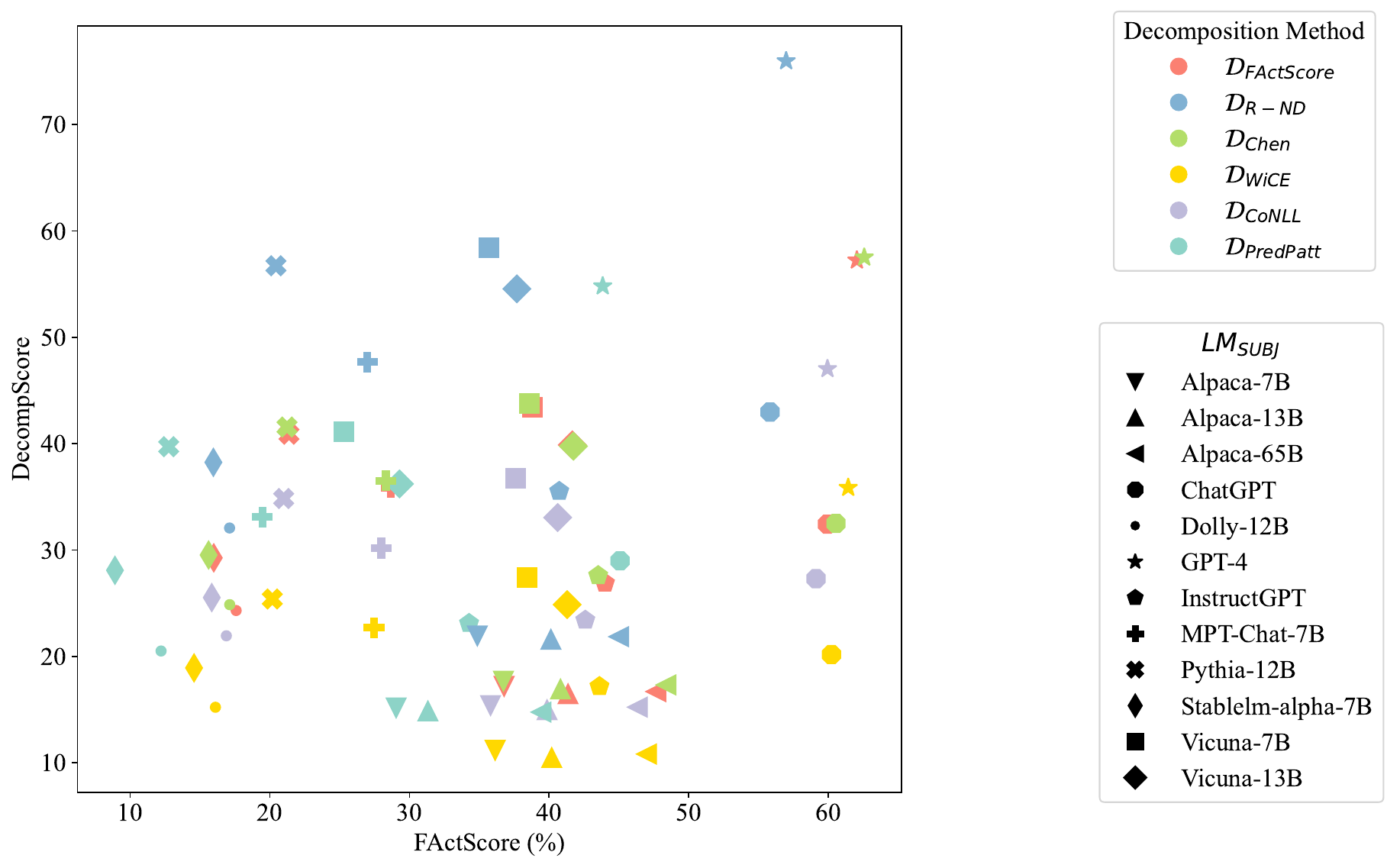}
    \caption{\factscore{} results after filtering out subclaims determined to be not supported by the original sentence (using \factscoresentencecontext{} judgments) for all claim decomposition methods and $\textrm{LM}_{\textrm{SUBJ}}$.}
    \label{fig:scatter_all_filtered_factscore}
\end{figure*}

\factscore{} results based on the subclaims judged to cohere with the original claim (based on judgments obtained when computing \factscoresentencecontext{}) are shown in \autoref{tab:filtered factscore} and \autoref{fig:scatter_all_filtered_factscore}. The average numbers of subclaims per biography are reported in \autoref{tab:subclaims}, and the average numbers of supported subclaims (i.e., the \factscoresentencecontext{}) are reported in \autoref{tab:filtered subclaims}.

It is important to note the special cases and conditions placed on these results: \begin{itemize}
    \item The released data from \citet{min2023factscore} includes invalid LM responses (e.g. “I'm sorry, I don't have any information on a person named…”). Including these generations is valuable for evaluating factuality of a language model, however results in noise when evaluating decomposition quality. These invalid responses are still processed by the decomposition methods we wish to evaluate, however the quality of decomposition is unaffected.
    \item Different language models are trained on different versions of Wikipedia, which introduces inconsistencies from the Wikipedia context used for fact-checking. This can affect \factscore{} but does not affect \factscoresentencecontext{} because it does not make use of external knowledge sources.%
\end{itemize}

\begin{table*}
\begin{center}
    \begin{tabular}{cccccccc}
        \hline
        \multicolumn{8}{c}{\factscoresentencecontext{}}\\
        \hline
        $\textrm{LM}_{\textrm{SUBJ}}$ & \prompt{\ourmethod{}} & \prompt{Chen} & \prompt{\wice{}} & \prompt{FS} & \prompt{FS2} & \prompt{CoNLL} & \prompt{PP} \\
        \hline\hline
        Alpaca 7B        & \best{21.9} & 17.7 & 11.2 & 17.2 & 18.8 & 15.4 & 15.2 \\
        Alpaca 13B       & \best{21.6} & 16.9 & 10.5 & 16.5 & 18.2 & 15.0 & 14.9 \\
        Alpaca 65B       & \best{21.9} & 17.3 & 10.8 & 16.7 & 18.5 & 15.2 & 14.8 \\
        ChatGPT          & \best{43.0} & 32.5 & 20.2 & 32.4 & 33.9 & 27.3 & 29.0 \\
        Dolly 12B        & \best{32.1} & 24.9 & 15.2 & 24.3 & 26.8 & 21.9 & 20.5 \\
        GPT4             & \best{76.0} & 57.5 & 35.9 & 57.2 & 58.5 & 47.0 & 54.8 \\
        InstructGPT      & \best{35.5} & 27.6 & 17.2 & 26.9 & 28.8 & 23.4 & 23.1 \\
        MPT-Chat 7B      & \best{47.7} & 36.5 & 22.7 & 35.9 & 37.4 & 30.2 & 33.1 \\
        Oasst-pythia 12B & \best{56.7} & 41.6 & 25.4 & 40.9 & 42.3 & 34.8 & 39.7 \\
        StableLM 7B      & \best{38.2} & 29.5 & 18.9 & 29.3 & 30.6 & 25.5 & 28.1 \\
        Vicuna 7B        & \best{58.4} & 43.8 & 27.4 & 43.4 & 45.4 & 36.7 & 41.1 \\
        Vicuna 13B       & \best{54.6} & 39.8 & 24.9 & 39.9 & 41.5 & 33.1 & 36.2 \\
        \hline
        Macro-average & \best{42.3} & 32.1 & 20.0 & 31.7 & 33.4 & 27.1 & 29.2 \\
    \hline
    \end{tabular}
    \caption{\factscoresentencecontext{} for each decomposition method and $\textrm{LM}_{\textrm{SUBJ}}$. Average number of subclaims generated per biography that are determined to be supported by the original sentence.}
    \label{tab:filtered subclaims}
\end{center}
\end{table*}

\begin{table*}
\begin{center}
    \begin{tabular}{cccccccc}
        \hline
        \multicolumn{8}{c}{\# Subclaims} \\
        \hline
        $\textrm{LM}_{\textrm{SUBJ}}$ & \prompt{\ourmethod{}} & \prompt{Chen} & \prompt{\wice{}} & \prompt{FS} & \prompt{FS2} & \prompt{CoNLL} & \prompt{PP} \\
        \hline\hline
        Alpaca 7B        & \best{22.2} & 17.9 & 11.3 & 17.3 & 19.0 & 15.7 & 16.4 \\
        Alpaca 13B       & \best{22.0} & 17.2 & 10.6 & 16.6 & 18.4 & 15.3 & 16.2 \\
        Alpaca 65B       & \best{22.2} & 17.5 & 10.9 & 16.9 & 18.6 & 15.5 & 16.0  \\
        ChatGPT          & \best{44.2} & 33.0 & 20.4 & 33.0 & 34.6 & 28.5 & 33.2 \\
        Dolly 12B        & \best{33.0} & 25.2 & 15.4 & 24.7 & 27.2 & 22.9 & 23.4 \\
        GPT4             & \best{77.7} & 58.2 & 36.2 & 57.9 & 59.2 & 48.6 & 63.6 \\
        InstructGPT      & \best{36.3} & 27.9 & 17.3 & 27.2 & 29.1 & 23.9 & 25.6 \\
        MPT-Chat 7B      & \best{49.0} & 37.0 & 22.9 & 36.3 & 37.8 & 31.1 & 37.4 \\
        Oasst-pythia 12B & \best{57.7} & 41.8 & 25.5 & 41.2 & 42.6 & 35.4 & 44.6 \\
        StableLM 7B      & \best{40.4} & 30.7 & 19.4 & 30.4 & 32.0 & 27.4 & 33.4 \\
        Vicuna 7B        & \best{59.8} & 44.3 & 27.6 & 43.9 & 45.9 & 37.7 & 46.3 \\
        Vicuna 13B       & \best{57.3} & 44.6 & 25.1 & 45.8 & 42.8 & 34.8 & 42.2 \\
        \hline
        Macro-average & \best{43.5} & 32.9 & 20.2 & 32.6 & 33.9 & 28.1 & 33.2 \\
    \hline
    \end{tabular}
    \caption{Average number of subclaims generated per biography.}
    \label{tab:subclaims}
\end{center}
\end{table*}

\begin{table*}
\begin{center}
    \begin{tabular}{cccccccc}
        \hline
        \multicolumn{8}{c}{\factscore{} (\%)}\\
        \hline
        $\textrm{LM}_{\textrm{SUBJ}}$ & \prompt{\ourmethod{}} & \prompt{Chen} & \prompt{\wice{}} & \prompt{FS} & \prompt{FS2} & \prompt{CoNLL} & \prompt{PP} \\
        \hline\hline
        Alpaca 7B        & 35.0 & 36.9 & 33.7 & 36.9 & 37.5 & 34.9 & 27.4 \\
        Alpaca 13B       & 38.9 & 40.3 & 35.1 & 40.8 & 41.1 & 38.3 & 30.0 \\
        Alpaca 65B       & 44.0 & 47.0 & 42.8 & 46.9 & 47.3 & 45.0 & 36.5 \\
        ChatGPT          & 48.2 & 52.1 & 51.4 & 52.2 & 52.2 & 50.7 & 36.8 \\
        Dolly 12B        & 16.5 & 16.3 & 13.9 & 16.7 & 17.2 & 15.5 & 10.4 \\
        GPT4             & 51.1 & 56.1 & 54.8 & 55.9 & 54.9 & 53.3 & 35.6 \\
        InstructGPT      & 40.1 & 43.2 & 43.2 & 43.6 & 43.4 & 41.7 & 31.5 \\
        MPT-Chat 7B      & 24.8 & 25.9 & 24.4 & 26.2 & 25.2 & 25.1 & 16.1 \\
        Oasst-pythia 12B & 20.1 & 20.8 & 19.2 & 21.2 & 21.1 & 20.5 & 11.7 \\
        StableLM 7B      & 13.8 & 13.1 & 11.6 & 13.5 & 13.4 & 13.3 & 8.2  \\
        Vicuna 7B        & 32.4 & 34.5 & 34.0 & 35.2 & 34.9 & 33.8 & 21.7 \\
        Vicuna 13B       & 31.1 & 32.8 & 31.8 & 34.1 & 35.7 & 33.1 & 23.3 \\
        \hline
        Macro-average & 33.0 & 34.9 & 33.0 & 35.3 & 35.3 & 33.8 & 24.1 \\
        \hline
    \end{tabular}
    \caption{\factscore{} of biographies generated by each $\textrm{LM}_{\textrm{SUBJ}}$ and decomposed with each method. Note: For evaluating decomposition quality, a larger \factscore{} is not necessarily better; we care about high confidence that \factscore{} is correct.}
    \label{tab:factscore}
\end{center}
\end{table*}

\begin{table*}
\begin{center}
    \begin{tabular}{cccccccc}
        \hline
        \multicolumn{8}{c}{\factscore{} (\%) After Filtering Out Unsupported Subclaims}\\
        \hline
        $\textrm{LM}_{\textrm{SUBJ}}$ & \prompt{\ourmethod{}} & \prompt{Chen} & \prompt{\wice{}} & \prompt{FS} & \prompt{FS2} & \prompt{CoNLL} & \prompt{PP} \\
        \hline\hline
        Alpaca 7B        & 34.9 & 36.7 & 36.1 & 36.8 & 37.6 & 35.8 & 29.1 \\
        Alpaca 13B       & 40.1 & 40.8 & 40.2 & 41.4 & 41.2 & 39.9 & 31.3 \\
        Alpaca 65B       & 45.0 & 48.4 & 47.0 & 47.6 & 47.9 & 46.3 & 39.4 \\
        ChatGPT          & 55.8 & 60.5 & 60.2 & 59.9 & 59.9 & 59.1 & 45.1 \\
        Dolly 12B        & 17.1 & 17.1 & 16.1 & 17.6 & 17.7 & 16.9 & 12.2 \\
        GPT4             & 57.0 & 62.6 & 61.4 & 62.0 & 61.0 & 59.9 & 43.8 \\
        InstructGPT      & 40.7 & 43.5 & 43.6 & 44.0 & 44.0 & 42.6 & 34.3 \\
        MPT-Chat 7B      & 27.0 & 28.3 & 27.5 & 28.7 & 27.6 & 28.0 & 19.5 \\
        Oasst-pythia 12B & 20.4 & 21.2 & 20.2 & 21.4 & 21.4 & 21.0 & 12.8 \\
        StableLM 7B      & 16.0 & 15.6 & 14.6 & 16.0 & 15.8 & 15.9 & 8.9  \\
        Vicuna 7B        & 35.7 & 38.6 & 38.4 & 38.8 & 38.4 & 37.6 & 25.3 \\
        Vicuna 13B       & 37.7 & 41.7 & 41.3 & 41.7 & 41.1 & 40.6 & 29.3 \\
        \hline
        Macro-average & 35.6 & 37.9 & 37.2 & 38.0 & 37.8 & 37.0 & 27.6 \\
    \hline
    \end{tabular}
    \caption{\factscore{} of biographies after filtering out subclaims determined to be not supported by the original sentence (using \factscoresentencecontext{} judgments). Note: For evaluating decomposition quality, a larger \factscore{} is not necessarily better; we care about high confidence that \factscore{} is correct.}
    \label{tab:filtered factscore}
\end{center}
\end{table*}

\begin{table*}
\begin{center}
    \begin{tabular}{cccccccc}
        \hline
        \multicolumn{8}{c}{\% Subclaims Supported}\\
        \hline
        $\textrm{LM}_{\textrm{SUBJ}}$ & \prompt{\ourmethod{}} & \prompt{Chen} & \prompt{\wice{}} & \prompt{FS} & \prompt{FS2} & \prompt{CoNLL} & \prompt{PP} \\
        \hline\hline
        Alpaca 7B        & 98.7 & 98.9 & \best{99.2} & 99.1 & 99.1 & 98.4 & 93.6 \\
        Alpaca 13B       & 98.6 & 99.0 & \best{99.4} & 99.0 & 99.2 & 98.2 & 93.2 \\
        Alpaca 65B       & 98.6 & 99.3 & \best{99.4} & 99.2 & 99.3 & 98.5 & 93.7 \\
        ChatGPT          & 93.0 & 95.9 & 96.7 & \best{99.4} & 94.5 & 89.0 & 80.0 \\
        Dolly 12B        & 97.4 & 98.7 & \best{99.0} & 98.7 & 98.6 & 96.5 & 89.6 \\
        GPT4             & 96.2 & 97.4 & \best{98.3} & 97.4 & 97.2 & 94.2 & 83.2 \\
        InstructGPT      & 98.1 & 99.1 & \best{99.3} & 99.0 & 99.0 & 98.0 & 90.8 \\
        MPT-Chat 7B      & 96.5 & 97.6 & \best{98.4} & 97.6 & 97.8 & 95.4 & 86.9 \\
        Oasst-pythia 12B & 98.3 & 99.3 & \best{99.4} & 99.3 & 99.3 & 98.4 & 89.4 \\
        StableLM 7B      & 89.2 & 90.7 & \best{94.1} & 90.5 & 89.4 & 84.8 & 74.4 \\
        Vicuna 7B        & 94.8 & 97.0 & \best{98.1} & 96.3 & 96.5 & 92.9 & 84.1 \\
        Vicuna 13B       & 88.9 & 93.3 & \best{95.4} & 90.8 & 88.1 & 82.6 & 72.6 \\
        \hline
        Macro-average & 96.0 & 97.2 & \best{98.1} & 97.2 & 96.5 & 93.9 & 86.0 \\
        \hline
    \end{tabular}
    \caption{Percentage of subclaims from each decomposition method and $\textrm{LM}_{\textrm{SUBJ}}$ that are judged to be supported by (cohere with) the original claim.}
    \label{tab:d-factscore}
\end{center}
\end{table*}

\section{Model Details}\label{sec:model specs}

To reduce cost using the \texttt{text-davinci-003} model used by \citet{min2023factscore}, we instead use InstructGPT (\texttt{gpt-3.5-turbo-instruct}) as the LLM for decomposition with 4K token context window, 512 \texttt{max\_tokens} and a temperature of 0.7. This model costs \$0.0015 per 1K input tokens and \$0.0020 per 1K output tokens. \texttt{gpt-3.5-turbo-instruct} achieves Pearson correlation coefficients of over 0.97 for both \factscore{} and number of subclaims generated compared to results reported by \citet{min2023factscore} (\autoref{tab:model correl}).

\begin{table*}
\begin{center}
    \begin{tabular}{ccccc}
        \hline
            & \factscore{}   & Reported \factscore{} & \# subclaims   & Reported \# subclaims \\
        \hline\hline
        Alpaca 7B                       & 36.9 & 36.5   & 17.3 & 17.4   \\
        Alpaca 13B                      & 40.8 & 40.3   & 16.6 & 16.6   \\
        Alpaca 65B                      & 46.9 & 46.3   & 16.9 & 17.1   \\
        ChatGPT                         & 52.2 & 60.4   & 33.0 & 37.0   \\
        Dolly 12B                       & 16.7 & 17.1   & 24.7 & 24.6   \\
        GPT4                            & 55.9 & 59.9   & 57.9 & 60.8   \\
        InstructGPT                     & 43.6 & 41.7   & 27.2 & 27.7   \\
        MPT-Chat 7B                     & 26.2 & 27.9   & 36.3 & 37.3   \\
        Oasst-pythia 12B                & 21.2 & 20.8   & 41.2 & 39.7   \\
        StableLM 7B                     & 13.5 & 16.3   & 30.4 & 38.0   \\
        Vicuna 7B                       & 35.2 & 36.9   & 43.9 & 45.6   \\
        Vicuna 13B                      & 34.1 & 40.7   & 45.8 & 50.9   \\
        \hline
        $\rho$         &    \multicolumn{2}{c}{0.9786} & \multicolumn{2}{c}{0.9821}  \\
        \hline
    \end{tabular}
    \caption{Pearson correlation coefficients ($\rho$) between our setup for computing \factscore{} (using \texttt{gpt-3.5-turbo-instruct} for subclaim generation) and results reported by \protect\citet{min2023factscore} (using \texttt{text-davinci-003} for subclaim generation).}
    \label{tab:model correl}
\end{center}
\end{table*}

Inst-LLAMA is \textsc{LLAMA} trained on Super Natural Instructions \citep{wang-etal-2022-super, touvron2023llama}, and is used for all \factscore{} and \factscoresentencecontext{} evaluations. We use \texttt{max\_sequence\_length} of 2048 and \texttt{max\_output\_length} of 128.

For \prompt{PredPatt}, we use \texttt{Trankit} for generating the dependency parse for each sentence. This parse is then used by \texttt{PredPatt} with the following flags: relative clauses, appositional modifiers, adjectival modifiers, conjunction, possessives, borrow\_arg\_for\_relcl and strip all set to True, with the remaining flags (simple, cut, and big\_args) set to False. We use \texttt{PredPatt} with Universal Dependencies v2.

We use \texttt{gpt-3.5-turbo-instruct} with the settings enumerated above for converting PredPatt outputs into natural language sentences with the following prompt:\\

{\small \noindent Please turn my input utterances into a grammatically correct natural English sentence by resolving tense, fixing grammatical errors, and reordering words without changing meanings. Your output should not contain ``is/are'' or ``poss''. Your output should contain no hallucinated information and no redundant sentences. Just the modified utterance.
\\

\noindent Input: born 1908 community leader

\noindent Output: The community leader was born in 1908.
\\

\noindent Input: date of death is/are unknown

\noindent Output: The date of death is unknown.
\\

\noindent Input: was an African - American social worker activist

\noindent Output: They were an African-American social worker activist.
\\

\noindent Input: <subclaim>

\noindent Output:}
\\

When a prompt in the \prompt{CoNLL-U} approach exceeds the length allowed for the context window, examples are incrementally removed until the prompt fits. When a zero-shot prompt (no in-context examples) exceeds the size of the context window, we backoff and set the entire original sentence as the subclaim. In practice, we backoff 0.05\% of the time: across 6000 passages (500 passages generated by each of 12  $\textrm{LM}_{\textrm{SUBJ}}$), twice we use one example and once we use the original sentence. We leave it to future work to reduce the size of the parses used in the prompt.

\section{NLI Entailment}
\label{app:nli_entailment}
\begin{figure}[h]
    \centering
    \includegraphics[width=0.4\textwidth]{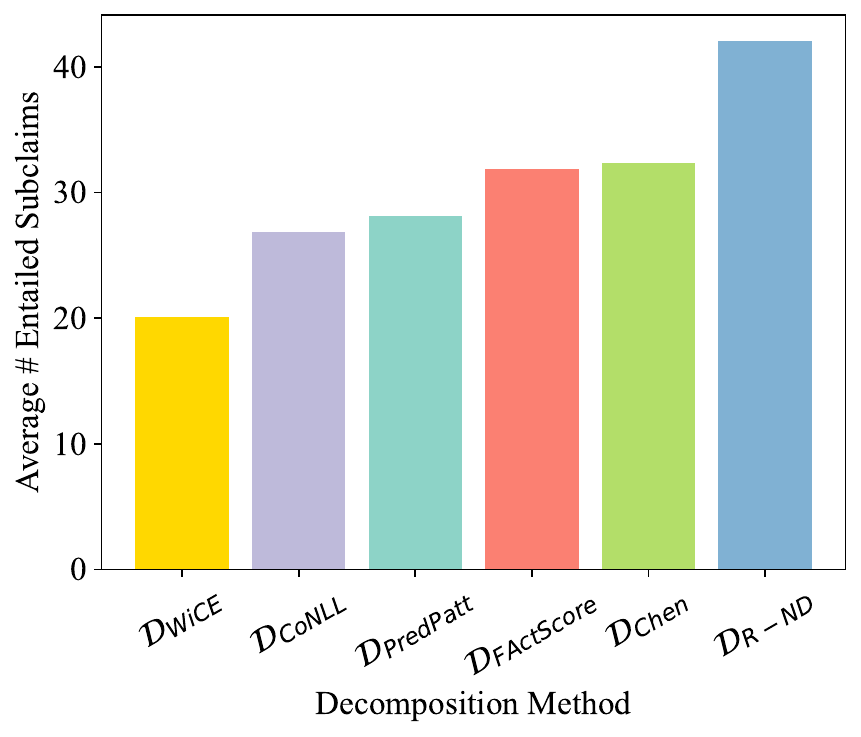}
    \caption{Average number of subclaims per passage that are entailed by their original sentential claim, as determined by an NLI model \protect\citep{nie-etal-2020-adversarial}. Values are macro-averaged across $\textrm{LM}_{\textrm{SUBJ}}$.}
    \label{fig:nli-entailment}
\end{figure}

The numbers of subclaims that are judged to be entailed by the original sentence are highly correlated with the numbers of subclaims judged by an LLM to be supported by the original sentence (\factscoresentencecontext{}), achieving a Pearson correlation coefficient of 0.9978 (\autoref{fig:nli-entailment}).

\section{Decomposition Examples}
\label{appendix:decomposition_outputs}
We include examples of two sentences decomposed manually and by all claim decomposition methods evaluated. The decompositions for the sentence ``Alfred Hitchcock passed away on April 29, 1980, in Bel-Air,
California, leaving behind a rich legacy of suspenseful and thrilling films that continue to captivate and inspire
audiences and filmmakers alike." are shown in \autoref{tab:hitchcock_decompositions}. The decompositions for the sentence ``Nash demonstrated a natural aptitude for mathematics from a
young age and earned his bachelor’s and master’s degrees in mathematics from the Carnegie Institute of Technology (now Carnegie Mellon University) in 1948.'' are shown in \autoref{tab:nash_decompositions}.

\section{Russellian/Neo-Davidsonian In-context Learning Examples}
The manually decomposed sentences used as in-context examples for \ourmethod{} are shown in \autoref{tab:russellian_examples}.

\begin{table*}
\centering
\small
    \small
    \begin{tabular}{cl}
        \hline
        Manual & - Alfred Hitchcock passed away. \\
                        & - Alfred Hitchcock's death occurred in April.\\
                        & - Alfred Hitchcock's death occurred on the 29th day of a month. \\
                        & - Alfred Hitchcock's death occurred in 1980. \\
                        & - Alfred Hitchcock's death occurred in Bel-Air. \\ 
                        & - Bel-Air is in California. \\ 
                        & - Alfred Hitchcock left behind a legacy. \\ 
                        & - Alfred Hitchcock’s legacy is rich. \\
                        & - Alfred Hitchcock’s legacy contains films. \\
                        & - The films in Alfred Hitchcock’s legacy are suspenseful. \\
                        & - The films in Alfred Hitchcock’s legacy are thrilling. \\
                        & - The films in Alfred Hitchcock’s legacy captivate audiences. \\
                        & - The films in Alfred Hitchcock’s legacy captivate filmmakers. \\
                        & - The films in Alfred Hitchcock’s legacy inspire audiences. \\
                        & - The films in Alfred Hitchcock’s legacy inspire filmmakers. \\
        \hline
        \ourmethod{}  & - Alfred Hitchcock passed away on April 29, 1980. \\
                        & - His death occurred in Bel-Air, California. \\
                        & - Alfred Hitchcock had a legacy. \\
                        & - Alfred Hitchcock’s legacy contains suspenseful films. \\
                        & - Alfred Hitchcock’s legacy contains thrilling films. \\
                        & - Alfred Hitchcock’s films continue to captivate audiences. \\
                        & - Alfred Hitchcock’s films continue to inspire filmmakers. \\
                        & - Alfred Hitchcock left behind his legacy. \\
        \hline
        \prompt{Chen et al.}     & - Alfred Hitchcock passed away on April 29, 1980. \\
                        & - Alfred Hitchcock passed away in Bel-Air, California. \\
                        & - Alfred Hitchcock left behind a rich legacy. \\
                        & - Alfred Hitchcock's legacy includes suspenseful and thrilling films. \\
                        & - Alfred Hitchcock's films continue to captivate and inspire audiences. \\
                        & - Alfred Hitchcock's films continue to captivate and inspire filmmakers. \\
        \hline
        \prompt{\wice{}}         & - Alfred Hitchcock passed away on April 29, 1980. \\
                        & - He passed away in Bel-Air, California. \\
                        & - He left behind a rich legacy of suspenseful and thrilling films. \\
                        & - His films continue to captivate and inspire audiences. \\
                        & - His films continue to inspire filmmakers. \\
        \hline
        \prompt{\factscore{}}       & - Alfred Hitchcock passed away on April 29, 1980. \\
                        & - Alfred Hitchcock passed away in Bel-Air, California. \\
                        & - Alfred Hitchcock left behind a legacy of films. \\
                        & - Alfred Hitchcock's films were suspenseful and thrilling. \\
                        & - Alfred Hitchcock's films continue to captivate audiences. \\
                        & - Alfred Hitchcock's films continue to inspire filmmakers.\\
        \hline
        \prompt{CoNLL-U}         & - Alfred Hitchcock passed away on April 29, 1980. \\
                        & - He passed away in Bel-Air, California. \\
                        & - He left behind a rich legacy. \\
                        & - His legacy includes suspenseful and thrilling films. \\
                        & - His films continue to captivate and inspire audiences and filmmakers. \\
        \hline
        \prompt{PredPatt}        & - Alfred Hitchcock passed away on April 29 , 1980 , in Bel - Air. \\
                        & - Bel - Air is/are in California \\
                        & - Alfred Hitchcock left behind a rich legacy of suspenseful and thrilling films that \\ & \hspace{0.2cm} continue to captivate and inspire audiences and filmmakers alike \\
                        & - The legacy of suspenseful and thrilling films that continue to captivate and inspire \\ & \hspace{0.2cm} audiences and filmmakers alike is/are rich \\
                        & - The films that continue to captivate and inspire audiences and filmmakers alike \\ & \hspace{0.2cm} are suspenseful \\
                        & - Suspenseful and thrilling films continue to captivate audiences. \\
                        & - Suspenseful and thrilling films continue to captivate filmmakers alike. \\
                        & - That continue to inspire. \\
        \hline
    \end{tabular}
    \caption{Decomposition comparison for the sentence ``Alfred Hitchcock passed away on April 29, 1980, in Bel-Air, California, leaving behind a rich legacy of suspenseful and thrilling films that continue to captivate and inspire audiences and filmmakers alike.''}
    \label{tab:hitchcock_decompositions}
\end{table*}

\begin{table*}
    \centering
    \small
    \begin{tabular}{cl}
        \hline
        Manual      & - Nash demonstrated an aptitude for mathematics. \\
                    & - Nash’s aptitude for mathematics was natural. \\
                    & - Nash’s demonstration of an aptitude for mathematics began at a young age. \\
                    & - Nash earned his bachelor’s degree. \\
                    & - Nash’s bachelor’s degree is in mathematics. \\
                    & - Nash’s bachelor’s degree is from the Carnegie Institute of Technology. \\
                    & - Nash’s bachelor’s degree was earned in 1948. \\
                    & - Nash earned his master’s degree. \\
                    & - Nash’s master’s degree is in mathematics. \\
                    & - Nash’s master’s degree is from the Carnegie Institute of Technology. \\
                    & - Nash’s master’s degree was earned in 1948. \\
                    & - The Carnegie Institute of Technology is now Carnegie Mellon University. \\
        \hline
        \ourmethod{}  & - Nash demonstrated a natural aptitude for mathematics. \\
                        & - His natural aptitude for mathematics was evident from a young age. \\
                        & - He earned a bachelor's degree in mathematics. \\
                        & - He earned a master's degree in mathematics. \\
                        & - He earned his degrees from the Carnegie Institute of Technology. \\
                        & - The Carnegie Institute of Technology is now known as Carnegie Mellon University. \\
                        & - He earned his degrees in 1948. \\
                        & - He earned his bachelor's degree in mathematics in 1948. \\
                        & - He earned his master's degree in mathematics in 1948. \\
        \hline
        \prompt{Chen et al.}     & - Nash demonstrated a natural aptitude for mathematics. \\
                        & - Nash demonstrated a natural aptitude for mathematics from a young age. \\
                        & - Nash earned his bachelor's degree in mathematics. \\
                        & - Nash earned his master's degree in mathematics. \\
                        & - Nash earned his bachelor's and master's degrees in mathematics. \\
                        & - Nash earned his bachelor's and master's degrees in mathematics from the Carnegie \\ & \hspace{0.2cm} Institute of Technology. \\
                        & - The Carnegie Institute of Technology is now Carnegie Mellon University. \\
                        & - Nash earned his bachelor's and master's degrees in mathematics from the Carnegie \\ & \hspace{0.2cm} Institute of Technology (now Carnegie Mellon University). \\
                        & - Nash earned his bachelor's and master's degrees in mathematics from the Carnegie \\ & \hspace{0.2cm} Institute of Technology (now Carnegie Mellon University) in 1948. \\
        \hline
        \prompt{\wice{}}         & - Nash demonstrated a natural aptitude for mathematics from a young age. \\
                        & - Nash earned his bachelor's degree in mathematics from the Carnegie Institute of \\ & \hspace{0.2cm} Technology in 1948. \\
                        & - Nash earned his master's degree in mathematics from the Carnegie Institute of \\ & \hspace{0.2cm} Technology in 1948. \\
                        & - The Carnegie Institute of Technology is now known as Carnegie Mellon University. \\
        \hline
        \prompt{\factscore{}}       & - Nash demonstrated a natural aptitude for mathematics. \\
                        & - Nash demonstrated a natural aptitude for mathematics from a young age. \\
                        & - Nash earned his bachelor's degree in mathematics. \\
                        & - Nash earned his master's degree in mathematics. \\
                        & - Nash earned his bachelor's and master's degrees in mathematics. \\
                        & - Nash earned his bachelor's and master's degrees in mathematics from the Carnegie \\ & \hspace{0.2cm} Institute of Technology. \\
                        & - Carnegie Institute of Technology is now called Carnegie Mellon University. \\
                        & - Nash earned his bachelor's and master's degrees in mathematics from the Carnegie \\ & \hspace{0.2cm} Institute of Technology (now Carnegie Mellon University). \\
                        & - Nash earned his bachelor's and master's degrees in mathematics from the Carnegie \\ & \hspace{0.2cm} Institute of Technology (now Carnegie Mellon University) in 1948. \\
        \hline
        \prompt{CoNLL-U}         & - Nash demonstrated an aptitude for mathematics. \\
                        & - Nash earned his bachelor's and master's degrees in mathematics. \\
                        & - Nash's degrees were from Carnegie Institute of Technology. \\
                        & - The institute is now known as Carnegie Mellon University. \\
                        & - Nash received his degrees in 1948. \\
        \hline
        \prompt{PredPatt}        & - Nash demonstrated a natural aptitude for mathematics from a young age. \\
                        & - Aptitude for mathematics is natural. \\
                        & - They were young. \\
                        & - Nash earned his bachelor 's and master 's degrees in mathematics from the Carnegie \\ & \hspace{0.2cm} Institute of Technology in 1948. \\
                        & - He had a bachelor 's and master 's degrees in mathematics. \\
                        & - The bachelor possessed a master's degree. \\
                        & - The Carnegie Institute of Technology is now Carnegie Mellon University. \\
        \hline
    \end{tabular}
    \caption{Decomposition comparison for the sentence ``Nash demonstrated a natural aptitude for mathematics from a young age and earned his bachelor's and master's degrees in mathematics from the Carnegie Institute of Technology (now Carnegie Mellon University) in 1948.''}
    \label{tab:nash_decompositions}
\end{table*}

\begin{table*}
    \centering
    \small
    \begin{tabular}{l}
        \hline
        He made his acting debut in the film The Moon is the Sun's Dream (1992), and continued to appear in small and \\supporting roles throughout the 1990s. \\
        - He has an acting debut. \\
        - He acted in a film. \\
        - His acting debut was in a film. \\
        - His acting debut was in The Moon is the Sun’s Dream. \\
        - He acted in The Moon is the Sun’s Dream. \\
        - The Moon is the Sun’s Dream is a film. \\
        - The Moon is the Sun’s Dream was released in 1992. \\
        - His acting debut occurred in 1992. \\
        - He appeared in small roles. \\
        - He appeared in supporting roles. \\
        - His small roles occurred throughout the 1990s. \\
        - His supporting roles occurred throughout the 1990s. \\
        - His appearance in small roles occurred after his acting debut. \\
        - His appearance in supporting roles occurred after his acting debut. \\
        \hline
        He is also a successful producer and engineer, having worked with a wide variety of artists, including Willie Nelson, \\Tim McGraw, and Taylor Swift. \\
        - He is a producer. \\
        - He is successful at being a producer. \\
        - He is an engineer. \\
        - He is successful at being an engineer. \\
        - He has worked with a wide variety of artists. \\
        - Willie Nelson is an artist. \\
        - He has worked with Willie Nelson. \\
        - Tim McGraw is an artist. \\
        - He has worked with Tim McGraw. \\
        - Taylor Swift is an artist. \\
        - He has worked with Taylor Swift. \\
        \hline
        In 1963, Collins became one of the third group of astronauts selected by NASA and he served as the back-up Command \\Module Pilot for the Gemini 7 mission. \\
        - NASA selected a third group of astronauts. \\
        - Collins belonged to the third group of astronauts. \\
        - Collins was selected by NASA. \\
        - Collins’s selection by NASA occurred in 1963. \\
        - The Gemini 7 mission has a back-up Command Module Pilot. \\
        - Collins’s role in the Gemini 7 mission was as the back-up Command Module Pilot. \\
        - Collins participated in the Gemini 7 mission. \\
        \hline
        In addition to his acting roles, Bateman has written and directed two short films and is currently in development \\on his feature debut. \\
        - Bateman has acting roles. \\
        - Bateman has written short films. \\
        - The number of short films Bateman has written is two. \\
        - Bateman has directed short films. \\
        - The number of short films Bateman has directed is two. \\
        - Bateman is currently in development on his feature debut. \\
        - The two short films were made before his feature debut. \\
        - His acting roles came before his feature debut. \\
        \hline
        Michael Collins (born October 31, 1930) is a retired American astronaut and test pilot who was the Command Module\\ Pilot for the Apollo 11 mission in 1969. \\
        - Michael Collins was born in October. \\
        - Michael Collins was born on the 31st day of a month. \\
        - Michael Collins was born in 1930. \\
        - Michael Collins is retired. \\
        - Michael Collins is American. \\
        - Michael Collins was an astronaut. \\
        - Michael Collins was a test pilot. \\
        - Michael Collins participated in the Apollo 11 mission. \\
        - Michael Collins’s participation in the Apollo 11 mission occurred in 1969. \\
        - The Apollo 11 mission was active in 1969. \\
        - The day of Michael Collins’s birth occurred before his year of participation in the Apollo 11 mission. \\
        - The Apollo 11 mission had a Command Module Pilot. \\
        - Michael Collins’s role in the Apollo 11 mission was as the Command Module Pilot. \\
        \hline
    \end{tabular}
\end{table*}
\clearpage
\newpage
\begin{table*}
    \centering
    \small    
    \begin{tabular}{l}
        \hline
        He was an American composer, conductor, and musical director. \\
        - He was American. \\
        - He was a composer. \\
        - He was a conductor. \\
        - He was a musical director. \\
        \hline
        She currently stars in the romantic comedy series, Love and Destiny, which premiered in 2019. \\
        - She stars in Love and Destiny. \\
        - Love and Destiny is a series. \\
        - Love and Destiny is a romantic comedy. \\
        - Love and Destiny premiered in 2019. \\
        \hline
        His music has been described as a mix of traditional Mexican and Latin American styles, as well as \\jazz, folk, and rock. \\
        - He has music. \\
        - His music has been described. \\
        - His music has been described as a mix of styles. \\
        - His music has been described as containing elements of traditional styles of music. \\
        - His music has been described as containing elements of Mexican style of music. \\
        - His music has been described as containing elements of Latin American style of music. \\
        - His music has been described as containing elements of jazz music. \\
        - His music has been described as containing elements of folk music. \\
        - His music has been described as containing elements of rock music. \\
        \hline
        He also serves as an ambassador for the charity Leonard Cheshire Disability. \\
        - He has a role in Leonard Cheshire Disability. \\
        - His role in Leonard Cheshire Disability is as an ambassador. \\
        - Leonard Cheshire Disability is a charity. \\
        \hline
        He began his career in Nashville in the late 1950s and has since released numerous albums, including a greatest hits \\collection in 1999. \\
        - He has a career. \\
        - His career began in Nashville. \\
        - His career began in the late 1950s. \\
        - He has released albums. \\
        - His released albums are numerous. \\
        - He released a collection. \\
        - His collection contains greatest hits. \\
        - His collection was released in 1999. \\
        - The release of his albums occurred after he began his career. \\
        \hline
        He has been performing since the age of 8, when he joined a band in his hometown of Guadalajara and has since \\gone on to record six studio albums and several singles of his own original material. \\
        - He has been performing. \\
        - He started performing at the age of 8. \\
        - He joined a band. \\
        - He joined a band at the age of 8. \\
        - His band was in Guadalajara. \\
        - His hometown is Guadalajara. \\
        - He has recorded studio albums. \\
        - The number of studio albums he has recorded is six. \\
        - He has recorded singles. \\
        - He has several singles. \\
        - His studio albums are his own original material. \\
        - His singles are his own original material. \\
        - His recording of studio albums occurred after he joined a band. \\
        - His recording of singles occurred after he joined a band. \\
        \hline
        She is also the former President of the Malaysian Chinese Association (MCA) from 2010 to 2013. \\
        - She had a role in the Malaysian Chinese Association. \\
        - Her role in the Malaysian Chinese Association was as its President. \\
        - Her tenure as President of the Malaysian Chinese Association started in 2010. \\
        - Her tenure as President of the Malaysian Chinese Association ended in 2013. \\
        - MCA is another name for the Malaysian Chinese Association. \\
        \hline
        During his professional career, McCoy played for the Broncos, the San Diego Chargers, the Minnesota Vikings, \\and the Jacksonville Jaguars. \\
        - McCoy had a professional career. \\
        - McCoy played for the Broncos. \\
        - McCoy played for the San Diego Chargers. \\
        - The Chargers are from San Diego. \\
        - McCoy played for the Minnesota Vikings. \\
        - The Vikings are from Minnesota. \\
        - McCoy played for the Jacksonville Jaguars. \\
        - The Jaguars are from Jacksonville. \\
        \hline
        \end{tabular}
\end{table*}
\clearpage
\newpage
\begin{table*}
    \centering
    \small    
    \begin{tabular}{l}
        \hline
        Miller has been described as the architect of Trump's controversial immigration policies, and has previously worked \\for Alabama Senator Jeff Sessions on immigration issues. \\
        - Miller has been described. \\
        - Miller has been described as an architect. \\
        - Miller has been described as an architect of Trump’s controversial immigration policies. \\
        - Trump has immigration policies. \\
        - Trump’s immigration policies are controversial. \\
        - Miller worked for Jeff Sessions. \\
        - Jeff Sessions is a Senator. \\
        - Jeff Sessions represents Alabama. \\
        - Miller worked on immigration issues. \\
        - Miller’s work for Jeff Sessions involved immigration issues. \\
        \hline
        Her work is often described as whimsical and dreamlike. \\
        - She has work. \\
        - Her work has been described. \\
        - Her work is described as whimsical. \\
        - Her work is described as dreamlike. \\
        - The description of her work as whimsical has occurred often. \\
        - The description of her work as dreamlike has occurred often. \\
        \hline
        He graduated from the United States Military Academy in 1952, and then went on to serve in the \\United States Air Force. \\
        - He graduated from the United States Military Academy. \\
        - His graduation from the United States Military Academy occurred in 1952. \\
        - He served in the United States Air Force. \\
        - His service in the United States Air Force occurred after his graduation from the United States Military Academy. \\
        \hline
        He is best known for his roles in the films Memories of Murder (2003), The Host (2006), (...) and Parasite (2019). \\
        - He had a role in Memories of Murder. \\
        - Memories of Murder is a film. \\
        - Memories of Murder was released in 2003. \\
        - He had a role in The Host. \\
        - The Host is a film. \\
        - The Host was released in 2006. \\
        - He had a role in Parasite. \\
        - Parasite is a film. \\
        - Parasite was released in 2009. \\
        - His role in Memories of Murder is one of his best known. \\
        - His role in The Host is one of his best known. \\
        - His role in Parasite is one of his best known. \\
        \hline
        Song Kang-ho was born in Gongju, South Korea in 1967. \\
        - Song Kang-ho was born. \\
        - Song Kang-ho’s birth occurred in Gongju. \\
        - Song Kang-ho’s birth occurred in South Korea. \\
        - Song Kang-ho’s birth occurred in 1967. \\
        - Gongju is in South Korea. \\
        \hline
        He studied theater at Chung-Ang University in Seoul. \\
        - He studied. \\
        - He studied theater. \\
        - He studied at Chung-Ang University. \\
        - His study of theater occurred at Chung-Ang University. \\
        - Chung-Ang University is located in Seoul. \\
        \hline
        His breakthrough came with the leading role in the acclaimed crime-drama film Memories of Murder in 2003. \\
        - He had a breakthrough. \\
        - His breakthrough was based on a leading role. \\
        - His breakthrough was based on his role in Memories of Murder. \\
        - His breakthrough occurred in 2003. \\
        - He had a leading role. \\
        - He had a leading role in Memories of Murder. \\
        - Memories of Murder is a film. \\
        - The genre of Memories of Murder is crime-drama. \\
        - Memories of Murder is acclaimed. \\
        - Memories of Murder was released in 2003. \\
        \hline
        This was followed by the monster movie The Host in 2006, which became the highest-grossing film in \\Korean history at the time. \\
        - This was followed by The Host. \\
        - The Host is a movie. \\
        - The Host was released in 2006. \\
        - The genre of The Host is monster movie. \\
        - The Host became the highest-grossing film in Korean history. \\
        \hline
    \end{tabular}
    \caption{Manually decomposed examples used for in-context examples by \ourmethod{}.}
\label{tab:russellian_examples}
\end{table*}

\end{document}